\documentclass[10pt,conference]{IEEEtran}
\IEEEoverridecommandlockouts

\usepackage{algorithm}
\usepackage[noend]{algpseudocode}
\usepackage{amsfonts,amsmath,amssymb}
\usepackage{booktabs}
\usepackage{doi}
\usepackage{duckuments}
\usepackage[T1]{fontenc}
\usepackage{graphicx}
\usepackage[utf8]{inputenc}
\usepackage{microtype}
\usepackage{multirow}
\usepackage{nicefrac}
\usepackage[numbers,compress,sort]{natbib}
\usepackage[autolanguage]{numprint} \npthousandsep{,}
\usepackage{subcaption}
\usepackage{soul}
\usepackage{url}
\usepackage{xcolor}
\usepackage{footnote}

\makesavenoteenv{tabular}
\makesavenoteenv{table}

\newif\ifcomment
\commenttrue    % Uncomment for blue/red comments.
\newif\iffigabbrv
\figabbrvtrue    % Uncomment for "Fig.".
\newcommand{\figtext}{\iffigabbrv Fig.\else Figure\fi}

\newif\ifeqabbrv
\eqabbrvtrue    % Uncomment for "Eq.".
\newcommand{\ourproblem}{\textsc{Dynamic-Maximum-Independent-Set}}

\title{\texorpdfstring{Unsupervised Learning of Local Updates for\\ Maximum Independent Set in Dynamic Graphs}{Unsupervised Learning of Local Updates for Maximum Independent Set in Dynamic Graphs}}
\author{
    \IEEEauthorblockN{Devendra Parkar\IEEEauthorrefmark{1}\IEEEauthorrefmark{2}\IEEEauthorrefmark{3}, Anya Chaturvedi\IEEEauthorrefmark{1}\IEEEauthorrefmark{3}, and Joshua J. Daymude\IEEEauthorrefmark{1}\IEEEauthorrefmark{2}}
    \IEEEcompsocitemizethanks{\IEEEcompsocthanksitem\IEEEauthorrefmark{3}These authors contributed equally to this work.}
    \IEEEauthorblockA{\IEEEauthorrefmark{1}School of Computing and Augmented Intelligence\\
    \IEEEauthorrefmark{2}Biodesign Center for Biocomputing, Security and Society\\
    Arizona State University, Tempe, AZ 85281\\
    Emails: \texttt{dparkar1@asu.edu}, \texttt{anya.chaturvedi@asu.edu}, \texttt{jdaymude@asu.edu}}
}

\begin{document}

\maketitle

\begin{abstract}
    We present the first unsupervised learning model for Maximum-Independent-Set (MaxIS) in dynamic graphs where edges change over time.
    Our method combines structural learning from graph neural networks (GNNs) with a learned distributed update mechanism that, given an edge addition or deletion event, modifies nodes' internal memories and infers their MaxIS membership in a single, parallel step.
    We evaluate our model against a mixed integer programming solver and a breadth of unsupervised and supervised learning models for combinatorial optimization on static graphs.
    Across dynamic graphs of \numprint{200}--\numprint{1000} nodes, our model achieves approximation ratios that are competitive with the state-of-the-art models while running 1.91--6.70x faster.
    When generalizing to graphs with \numprint{100}x more nodes than those used for training, our model produces MaxIS solutions 1.00--1.18x larger than all other unsupervised models, but is outperformed by the state-of-the-art supervised model.
    These results demonstrate that this novel, unsupervised, update-based learning approach to dynamic combinatorial optimization is a viable alternative to the na\"ive reapplication of analogous models for static graphs, leveraging temporal information to improve neural methods for combinatorial optimization.
\end{abstract}

\section{Introduction} \label{sec:intro}

Combinatorial optimization problems on graphs (e.g., \textsc{Graph-Coloring}, \textsc{Traveling-Salesman}, \textsc{Vehicle-Routing}, etc.) arise naturally in many practical domains~\cite{Applegate2007-travelingsalesman,Toth2014-vehiclerouting}.
However, many of these problems are classically intractable to solve exactly, and some even resist efficient approximation.
The situation is exacerbated by the fact that many real-world applications are \textit{dynamic}, requiring algorithms to not only find solutions but also maintain them as the underlying structure evolves.
While one could iteratively apply algorithms designed for static graphs at each time step, this approach is often computationally expensive and inefficient.
Critically, this approach ignores the often close relationship between successive graph structures.
A more effective strategy might avoid repeatedly recomputing solutions from scratch by instead leveraging these small structural changes to incrementally update an existing solution.
Such \textit{update algorithms} are an active area of theoretical algorithms research~\cite{Assadi2019-fullydynamic,Hanauer2022-recentadvances,Chechik2019-fullydynamic,Behnezhad2019-fullydynamic,Zheng2019-computingnearmaximum,Gao2022-dynamicapproximate}, but no learning algorithms have yet taken this approach.

In this paper, we consider the \textsc{Maximum-Independent-Set} (MaxIS) problem in a dynamic setting.
For static graphs, MaxIS is classically known to be NP-hard~\cite{Garey1979-computersintractability} and hard to approximate: for general graphs, deterministic approximation within a constant factor is impossible~\cite{Robson1986-algorithmsmaximum} and within an $n^{1-\varepsilon}$ factor is NP-hard~\cite{Hastad1996-cliquehard}.
Nevertheless, MaxIS has a diverse range of real-world applications, from identifying functional nodes in brain networks~\cite{Afek2011-biologicalsolution} to constructing diversified investment portfolios~\cite{Hidaka2023-correlationdiversifiedportfolio}.
For example, different stocks can be modeled as nodes and those that are closely related have an edge between them to form a conflict graph; thus, the problem of building a diverse portfolio reduces to finding a high-value independent set.
In many of these situations, the conflict graph changes dynamically over time due to factors such as neuronal development, stock market fluctuations, and other unpredictable events.
Our goal is to solve MaxIS on such dynamically evolving graphs.

Recently, the learning community has developed several methods for solving combinatorial optimization problems on static graphs~\cite{Bengio2021-machinelearning}.
Central to many of these methods are Graph Neural Networks (GNNs) due to their efficiency in capturing the structural information present in real-world networks such as the Internet, social networks, and molecular interactions~\cite{Zhang2019-graphconvolutional,Wu2020-comprehensivesurvey,Liu2026-weightedgraph}.
In the dynamic setting, recent models such as temporal GNNs are primarily concerned with tasks such as edge prediction, node classification, graph classification, etc.~\cite{Xu2020-inductiverepresentation,Rossi2020-temporalgraph}.
Work considering combinatorial optimization problems in a dynamic setting remains limited, with few examples such as \citet{Gunarathna2022-dynamicgraph}, which is an indirect heuristic learning approach.
No existing models learn update mechanisms comparable to traditional update algorithms.

We propose a model that solves MaxIS in dynamic graphs by directly learning an update mechanism instead of a heuristic.
We utilize the power of message-passing GNNs~\cite{Battaglia2018-relationalinductive} combined with sequence learning to learn an update mechanism that maintains an approximate MaxIS solution as the underlying graph changes.
GNNs learn to aggregate structural information and update nodes' memberships in the MaxIS solution at each time step, while sequence learning extrapolates the effects of edge dynamics to changes in the MaxIS solution.
Having learned an update mechanism, the model is able to update a MaxIS solution following a change in the dynamic graph in a single inference step, unlike more popular heuristic learning methods which require multiple inference steps and thus scale poorly with graph size.
We underscore this as a necessary shift in perspective required to address some of the common criticisms of learning approaches to solving combinatorial optimization problems, namely, lack of scalability, computation inefficiency, and huge training sample requirements.

\subsubsection*{Contributions}

We present the first unsupervised learning model for \textsc{Maximum-Independent-Set} (MaxIS) in dynamic graphs.
The central idea of learning local, distributed update mechanisms for dynamic combinatorial optimization problems is novel, and we demonstrate its efficacy in our evaluations.
Compared to the state-of-the-art models for combinatorial optimization in static graphs (OptGNN~\cite{Yau2024-aregraph}, unsupervised; and Fast T2T~\cite{Li2024-fastt2t}, supervised), our model maintains competitive approximation ratios across a variety of dynamic graph topologies.
This ensemble of methods all match or outperform a commercial mixed integer programming solver~\cite{GurobiOptimizationLLC2024-gurobioptimizer} in a small fraction of its runtime, surpassing the abilities of earlier unsupervised models for combinatorial optimization~\cite{Brusca2023-maximumindependent,Karalias2020-erdosgoes,Ahn2020-learningwhat}.
One specific parameterization of our model, which strictly bounds message-passing and update learning in local neighborhoods around edge dynamics, is able to maintain this strong performance while running 1.91--6.70x faster than OptGNN or Fast T2T, showcasing the scalability of our approach.
In generalization experiments considering graphs 100x the size of those used for training, our model produces MaxIS solutions 1.00--1.18x larger than the other unsupervised models; however, both OptGNN and Fast T2T achieve better performance--runtime tradeoffs.

\section{Related Work} \label{sec:relwork}

\subsubsection*{Traditional Algorithms for MaxIS}

\textsc{Maximum-Independent-Set} (MaxIS) is a classical NP-complete problem in graph theory and has been explored since the 1970s in the field of combinatorial optimization.
Algorithms seeking exact solutions use reduction rules~\cite{Tarjan1977-findingmaximum}, branching heuristics~\cite{Fomin2013-exactexponential}, and kernelization techniques~\cite{Lamm2016-findingnearoptimal} to reduce the size of the search space.
Among exact algorithms, the state-of-the-art is \citet{Xiao2017-exactalgorithms} which runs in $1.1996^nn^{O(1)}$ time and polynomial space, where $n$ is the number of nodes in the graph.
Extensions of MaxIS to dynamic graphs remain relatively unexplored.
\citet{Zheng2019-computingnearmaximum} and \citet{Gao2022-dynamicapproximate} have developed update algorithms using rule-based optimization techniques for incrementally maintaining an existing solution in response to edge insertions or deletions.
However, no such approaches have been taken by the learning community, which we discuss next.

\subsubsection*{Learning in Dynamic Graphs}

Memory-based models utilize various forms of persistent, high-dimensional representations (i.e., memories) for nodes to learn the dynamics of a graph as it changes.
These have been very successful in efficiently learning prediction and classification tasks in dynamic graphs~\cite{Rossi2020-temporalgraph,Kumar2019-predictingdynamic,Trivedi2019-dyreplearning,Yu2023-betterdynamic}.
Some models also include a GNN-based graph embedding module to mitigate the problem of memory staleness, i.e., when a node's memory hasn't been updated for a long time.
We utilize these innovations in our model.
Other learning models for dynamic graphs do not involve memory~\cite{Xu2020-inductiverepresentation,Pareja2020-evolvegcnevolving,Longa2023-graphneural} but have never been applied to combinatorial optimization problems.

\subsubsection*{Learning for Combinatorial Optimization}

A vast body of machine learning research considers combinatorial optimization problems~\cite{Peng2021-graphlearning,Dai2017-learningcombinatorial,Bengio2021-machinelearning}.
Among these, GNNs are the most common backbone given their excellent ability to characterize the structural information of graphs~\cite{Zhang2019-graphconvolutional,Wu2020-comprehensivesurvey}.
For MaxIS specifically, autoregressive approaches (i.e., those that take multiple inference steps to output a solution) such as reinforcement learning (RL) and dynamic programming (DP) have emerged as natural choices due to their easy constraint enforcement and a priori integral solution output~\cite{Ahn2020-learningwhat,Brusca2023-maximumindependent,Zhang2023-letflows}.
Yet, these methods suffer from sample inefficiency, requiring a number of inference steps that scales linearly with problem size~\cite{Karalias2020-erdosgoes,Peng2021-graphlearning}.
Non-autoregressive approaches (i.e., "single-shot" methods) have also been proposed, including some supervised, diffusion-based models~\cite{Sun2023-difuscographbased,Li2024-fastt2t} and a broader body of work on unsupervised, GNN-based methods~\cite{Li2018-combinatorialoptimization,Karalias2020-erdosgoes,Wang2022-unsupervisedlearning,Schuetz2022-combinatorialoptimization,Yau2024-aregraph}.
However, these only apply to static graphs.
To our knowledge, there is only one other learning model that explicitly considers dynamic graphs~\cite{Gunarathna2022-dynamicgraph}, and it applies RL to dynamic versions of the \textsc{Traveling-Salesman} and \textsc{Vehicle-Routing} problems.
Ours is the first non-autoregressive, unsupervised learning method for a combinatorial optimization problem in dynamic graphs.

\section{Maximum Independent Sets in Dynamic Graphs} \label{sec:prelim}

Consider an undirected, static graph $G = (V, E)$ with nodes $V$ and edges $E$.
An \textit{independent set} $I \subseteq V$ of $G$ is a set of nodes no two of which are adjacent, i.e., for all $u, v \in I$, we have $(u, v) \not\in E$.
Its \textit{size} is its number of nodes, denoted $|I|$.
A \textit{maximum independent set} (MaxIS) of $G$ is any independent set $I$ of $G$ satisfying $|I| = \max\{|I'| : I' \text{ is an independent set of } G\}$.

\textit{Dynamic graphs} (also called dynamic networks, temporal graphs, or evolving graphs) are graphs whose nodes and/or edges change over time~\cite{Bui-Xuan2003-computingshortest,Casteigts2012-timevaryinggraphs,Rossi2020-temporalgraph,Kazemi2020-representationlearning}.
For this work, we consider dynamic graphs where at most one edge changes per time and the underlying set of nodes is fixed.
Formally, a dynamic graph $\mathcal{G}$ is a finite sequence of \textit{graph snapshots} $(G_0, G_1, \ldots, G_T)$ where each snapshot $G_t = (V, E_t)$ is an undirected, static graph on nodes $V$ and, for all time steps $0 < t \leq T$, we have $|E_{t-1} \oplus E_t| = 1$, i.e., there is exactly one edge addition or one edge deletion per time step.
We denote the \textit{edge event} at time $t$ which transitions $G_{t-1}$ to $G_t$ as $\mathcal{E}_t$.
The \textit{neighborhood} of a node $v$ at time $t$ is $\mathcal{N}_t(v) = \{u \in V \mid (u,v) \in E_t\}$ and its \textit{degree} is denoted $d_t(v) = |\mathcal{N}_t(v)|$.
The \textit{$x$-neighborhood} of an edge event $\mathcal{E}_t$ are all nodes $u$ such that the shortest path in $G_t$ from $u$ to an endpoint of the edge being added or deleted in $\mathcal{E}_t$ contains at most $x$ edges.
The diameter of graph snapshot $G_t$, denoted $\text{diam}(G_t)$, is the maximum shortest hop-distance between any two nodes in $G_t$.

We define the \ourproblem\ (DynMaxIS) problem as follows:
Given a dynamic graph $\mathcal{G} = (G_0, G_1, \ldots, G_T)$---or, equivalently, an initial graph snapshot $G_0$ and a sequence of edge events $(\mathcal{E}_1, \ldots, \mathcal{E}_T)$---obtain a MaxIS for each snapshot $G_t$.
This problem is clearly NP-hard since it contains the NP-complete MaxIS problem as a special case ($T = 0$), so we are realistically interested in obtaining a large \textit{approximate} MaxIS for each snapshot.

\section{Unsupervised Learning Model for DynMaxIS} \label{sec:model}

As discussed in Section \ref{sec:relwork}, memory-based models have been quite successful in recent years.
Motivated by their effectiveness in predictive learning of dynamics, we present an unsupervised learning model focused on the MaxIS problem in dynamic graphs.
Specifically, we repurpose the two key modules of Temporal Graph Networks (TGN)~\cite{Rossi2020-temporalgraph}, memory and graph embedding, to learn an update mechanism that maintains a candidate MaxIS solution as the underlying graph changes.
Moreover, inspired by the use of messages (seen as random perturbations) to improve GNN performance for NP-hard problems~\cite{Papp2021-dropgnnrandom}, we couple node memory to an event handling module to inform nodes about nearby edge events.

\begin{figure*}[t]
    \centering
    \includegraphics[width=0.9\textwidth]{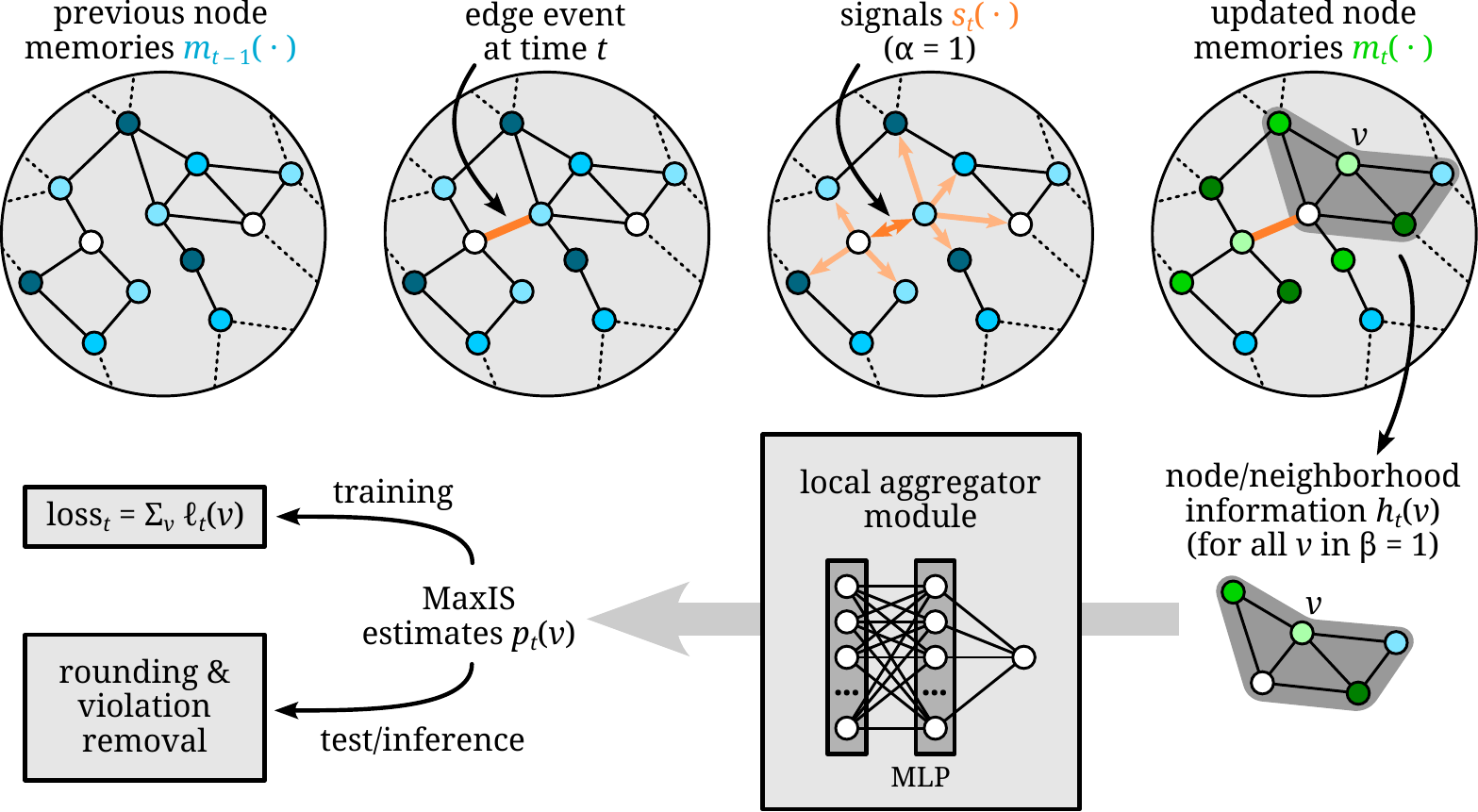}
    \caption{An overview of our unsupervised learning model for MaxIS on dynamic graphs.
    Nodes maintain memories (node colors) which are updated by signals propagating in the $\alpha$-neighborhoods of an edge event (orange arrows).
    Any node in the $\beta$-neighborhood of the edge event then locally aggregate its updated memory and those of its neighbors to estimate its new probability of MaxIS membership.
    These estimates are then used to compute loss (during training) or to generate an integral MaxIS solution (during testing and inference).}
    \label{fig:model}
\end{figure*}

Our model responds to edge events (i.e., edge additions or deletions) over time (see \figtext~\ref{fig:model}).
At each time step, the \textit{event handling module} is responsible for signaling all nodes within the edge event's $\alpha$-neighborhood that the event occurred.
These nodes then use the \textit{memory module} to update their internal representations using this signal.
Finally, nodes within the edge event's $\beta$-neighborhood use the \textit{local aggregation (graph embedding) module} to update their probabilities of membership in the MaxIS solution (i.e., an \textit{estimate}) based on their memory, degree, and memories of their neighbors.
% We describe each module and the training-related details of our model in the following subsections; pseudocode can be found in Appendix~\ref{app:pseudocode}.

\subsubsection*{Event Handling Module}

This module provides each node $v$ in the $\alpha$-neighborhood of the edge event $\mathcal{E}_t$ with a signal $s_t(v)$ encoding the type of event (addition or deletion) and the node's normalized distance from the event.
Formally, the signal for node $v$ at time $t$ is
\begin{equation} \label{eq:signal}
    s_t(v) = [\texttt{enc}(\mathcal{E}_t)~||~r_t(v)],
\end{equation}
where \texttt{enc} encodes edge deletions as $[0, 1]$ and edge additions as $[1, 0]$, and $r_t(v)$ is the hop-distance from $v$ to $\mathcal{E}_t$ normalized by $\alpha$ and linearly interpolated into $[-1, 1]$.

\subsubsection*{Memory Module}

A node's memory is a high-dimensional representation maintained throughout the model's execution that captures relevant structural changes affecting the node.
At each time $t$, this module updates the memories of nodes in the $\alpha$-neighborhood of the edge event $\mathcal{E}_t$.
Updates are performed using a GRU cell~\cite{Chung2014-empiricalevaluation}.
The memory of a node $v$ at time $t$ is
\begin{equation} \label{eq:memory}
    m_t(v) = \left\{\begin{array}{cl}
        \texttt{GRU}\big([m_{t-1}(v)~||~s_t(v)]) & \text{if $v$ in $\alpha$-nbrhd.}; \\
        m_{t-1}(v) & \text{otherwise}.
    \end{array}\right.
\end{equation}

\subsubsection*{Local Aggregation Module}

This module updates the estimates---i.e., probabilities of MaxIS membership---of all nodes in the $\beta$-neighborhood of the edge event $\mathcal{E}_t$.
First, each node $v$ aggregates its immediate neighbors' memories:
\begin{equation} \label{eq:aggnbrs}
    \tilde{h}_t(v) = \texttt{ReLU}\left(\sum_{u \in \mathcal{N}_t(v)}\mathbf{W}_1 m_t(u)\right),
\end{equation}
where $\mathbf{W}_1$ are learnable weights.
Next, node $v$ combines these aggregated memories with its own memory and degree information to obtain a local embedding
\begin{equation} \label{eq:aggself}
    h_t(v) = \mathbf{W}_2\left[\tilde{h}_t(v)~||~m_t(v)~||~d_t(v)\right],
\end{equation}
where again $\mathbf{W}_2$ are learnable weights.
Finally, node $v$ passes its embedding through a step-down MLP and a sigmoid function $\sigma$ to obtain its estimate in $[0, 1]$:
\begin{equation} \label{eq:estimate}
    p_t(v) = \sigma\big(\texttt{MLP}(h_t(v))\big).
\end{equation}
Any node $v$ beyond the edge event's $\beta$-neighborhood simply retains its previous estimate, i.e., $p_t(v) = p_{t-1}(v)$.

\subsubsection*{Update Training}

We first define a loss function from a node's local perspective and then define a cumulative loss function from these local losses.
This loss function is similar to those of other single-shot unsupervised methods, but enabling its calculation by individual nodes allows the memory and local aggregation modules to synergize, learning a distributed update mechanism.
The loss for a node $v$ at time $t$ is
\begin{equation} \label{eq:loss}
    \ell_t(v) = -p_t(v) + \frac{c}{2d_t(v)} \sum_{u \in \mathcal{N}_t(v)} p_t(u)p_t(v).
\end{equation}
The $-p_t(v)$ term rewards nodes with large estimates (i.e., probabilities of MaxIS membership closer to 1), aligning with the goal of finding a large independent set.
This is counterbalanced by the $\sum_{u \in \mathcal{N}_t(v)} p_t(u)p_t(v)$ term which penalizes violations of independence, i.e., when neighbors $u$ and $v$ both have large estimates.
This constraint violation term is normalized by the degree of $v$ and is then halved to compensate for double-counting the loss from the perspectives of both $u$ and $v$.
Finally, the constant $c$ is a hyperparameter balancing the two terms' loss contributions; in practice, $c = 3$ worked reasonably well across our evaluations (Section~\ref{sec:experiments}).

The cumulative loss at time $t$ is the sum of all local losses $\ell_t(v)$ for nodes $v$ in the edge event's $\beta$-neighborhood (i.e., all nodes that updated their estimates at time $t$).
This set of nodes at this one time step forms a single batch in our training process.
By setting $\beta$ as an appropriate function of dynamic graph size (see Model Variants, below), batch sizes will automatically scale to any input dynamic graph and provide sufficient inference samples for a training step.
The sequential processing of all edge events in a dynamic graph's training set forms a single epoch in our training process.
In a single run, we train until the losses for epochs have stabilized and select the epoch with the least loss.

\subsubsection*{Model Variants}

The neighborhood sizes $\alpha$ and $\beta$ control which nodes utilize an edge event to update their memories and/or update their estimates and contribute to cumulative loss, respectively.
In the \underbar{\textit{bounded cascading} (BCAS)} model variant, we set $\alpha = \beta = \gamma \cdot \text{diam}(G_0)$, where $\gamma \ll 1$, strictly bounding nodes' memory and estimate updates to a local region around each edge event.
This variant mimics non-learning-based update algorithms for MaxIS which analogously spread out from the point of the edge event, updating nodes' memberships to generate a new MaxIS~\cite{Behnezhad2019-fullydynamic,Chechik2019-fullydynamic}.
Locally bounding these update regions typically yields very fast update operations, at the potential risk of missing better MaxIS solutions that require far-reaching updates.

In the \underbar{\textit{non-cascading} (No-CAS)} model variant, $\alpha = 0$ and $\beta = \text{diam}(G_0)$; i.e., only the two nodes directly involved in the edge event update their memories, but all nodes in the graph update their estimates and contribute to cumulative loss.
In this variant, the local aggregation module is primarily responsible for the graph-wide learning of the update mechanism, with node memory aiding in the locality of the edge event.
This variant is important for dense graph topologies and edge dynamics where frequent update cascades may disrupt relevant information in nodes' memories.

\subsubsection*{Initial MaxIS Generation and Unsupervised Learning}

Update algorithms for MaxIS crucially require an existing MaxIS solution for the initial snapshot $G_0$ that is updated as the graph structure changes.
We mitigate this (potentially very expensive) prerequisite by utilizing an initial MaxIS generation phase that efficiently produces memories for all nodes in $G_0$ and model weights that training can warm-start from.
The idea is to build up $G_0$ one edge at a time (as a sequence of edge addition events), updating only the memories of the nodes involved in each edge addition.
Once $G_0$ is fully constructed, estimates and corresponding losses are computed for all nodes.
This constitutes a single epoch of this generation phase; in a single run of the generation phase, we execute epochs until the cumulative loss has stabilized.
After executing multiple runs with random seeds, we extract the node memories and model weights from the run with the least cumulative loss to warm-start model training.
Thus, the model as a whole learns to first find a candidate MaxIS (generation phase) and then maintain it (training phase) in an entirely unsupervised manner.

\subsubsection*{Integral Solution Generation} 

During testing and deployment, we use a simple rounding and violation removal procedure to convert the model's relaxed node estimates into an approximate MaxIS.
For a given snapshot $G_t$, all nodes $v$ with $p_t(v) \geq 0.5$ are initially included in the candidate solution.\footnote{Almost all estimates $p_t(v)$ are nearly-zero or nearly-one, so the threshold for inclusion could vary anywhere in $[0.25, 0.75]$ without changing the results. See Appendix~\ref{app:estimates} for details.}
Then, the following two steps are repeated until independence is achieved: (1) compute the number of \textit{violations} for each node in the candidate solution, i.e., the number of neighboring nodes that are also in the candidate solution; (2) remove any node with the most violations.
Such a procedure is a necessary part of single-shot learning methods that directly use relaxation to solve optimization problems with hard constraints~\cite{Karalias2020-erdosgoes,Donti2021-dc3learning,Schuetz2022-combinatorialoptimization,Wang2022-unsupervisedlearning}.

\section{Experiments} \label{sec:experiments}

% In this section, we describe our experimental methodology and then present and interpret the model evaluation results.

\subsection{Experimental Methodology} \label{subsec:methods}

\subsubsection*{Datasets}

We evaluate model performance using various initial graph structures with synthetic dynamics.
For initial graph structures, we consider RB-200~\cite{Wang2023-unsupervisedlearning}, which has been treated as the challenging evaluation instance for many recent combinatorial optimization learning methods~\cite{Brusca2023-maximumindependent,Li2024-fastt2t,Yau2024-aregraph}; BRAIN, a biological network with a sparse small-world topology~\cite{Crossley2013-cognitiverelevance}; and randomly generated Erd\H{o}s--R\'{e}nyi (ER) and Power Law (PL) graphs with \numprint{1000} nodes each.
Across all graphs, dynamics are produced by sampling edge additions or deletions from the initial snapshot's degree distribution\footnote{A na\"ive alternative is to sample edges uniformly at random, but this converges to homogeneous random snapshots over $n$ nodes after $n^2$ rounds, in expectation.
This makes studying different datasets uninteresting.} for a desired number of time steps; here, we use \numprint{50000} time steps for RB-200 and \numprint{100000} for the others.\footnote{These datasets' snapshots are divided chronologically by time step into 50:25:25 training, validation, and testing splits (70:15:15 for RB-200).}

\subsubsection*{Our Model and Comparisons}

We ran our complete training pipeline using the two model variants: BCAS and No-CAS.
For the BCAS variant, we set $\alpha = \beta = 0.25 \cdot \text{diam}(G_0)$, i.e., half of the radius of the first snapshot of the dynamic graph.
For each variant, we performed three training runs with different random seeds and report results for the median run.
Since ours is the only learning method for MaxIS in the dynamic setting, we compare our method against a commercial mixed integer programming solver and existing learning methods---both supervised and unsupervised---for MaxIS on static graphs applied to each snapshot of a dynamic graph independently.
\begin{itemize}
    \item \underbar{Gurobi}~\cite{GurobiOptimizationLLC2024-gurobioptimizer} is the most commonly used and commercially available mixed integer programming solver.
    We use Gurobi’s default settings for mixed integer program solving, which according to their documentation, is a branch-and-bound search.

    \item \underbar{OptGNN}~\cite{Yau2024-aregraph} is the state-of-the-art GNN-based, single-shot, unsupervised model for combinatorial optimization in static graphs.
    It leverages a novel semidefinite programming formulation to efficiently approximate various problems, which we extended to address MaxIS.

    \item \underbar{Erd\H{o}s-GNN}~\cite{Karalias2020-erdosgoes} is an earlier GNN-based, single-shot, unsupervised model for combinatorial optimization in static graphs.
    This also required new extensions for MaxIS.
    
    \item \underbar{DP-GNN}~\cite{Brusca2023-maximumindependent} combines dynamic programming with GNNs to form a multi-step, autoregressive comparator model.
    
    \item \underbar{LwDMIS}~\cite{Ahn2020-learningwhat} is a deep reinforcement learning method that adaptively adjusts its stages and defers decisions about nodes' MaxIS membership to improve scalability.

    \item \underbar{Fast T2T}~\cite{Li2024-fastt2t} is the state-of-the-art supervised diffusion model for combinatorial optimization in static graphs.
\end{itemize}

All methods were trained (if applicable) and tested on a machine with an AMD EPYC 7413 CPU and a 20 GB slice of one NVIDIA A100 GPU.
For the learning methods, training was halted after seven days.
For Gurobi, we enforced a time limit of 1~s per snapshot; this produces exact solutions on PL-1000, but yields approximations for the other datasets.

\subsubsection*{Metrics}

We evaluate each method using three metrics: performance, runtime, and peak memory usage.
Performance refers to the mean ratio of a method's candidate MaxIS size against that of Gurobi's across all snapshots in a dynamic graph.\footnote{Fast T2T does not specify a procedure for removing violations of independence, so we report the size of its candidate MaxIS as the number of nodes in its solution minus the number of pairwise violations.}
Runtime refers to the mean time taken by a method to process a single snapshot and generate a candidate MaxIS across all snapshots in a dynamic graph (reported as seconds per graph, s/g).
Peak memory usage denotes the maximum memory used by a method during training.

\subsection{Results} \label{subsec:results}

\begin{figure}[tb]
    \centering
    \includegraphics[width=\columnwidth]{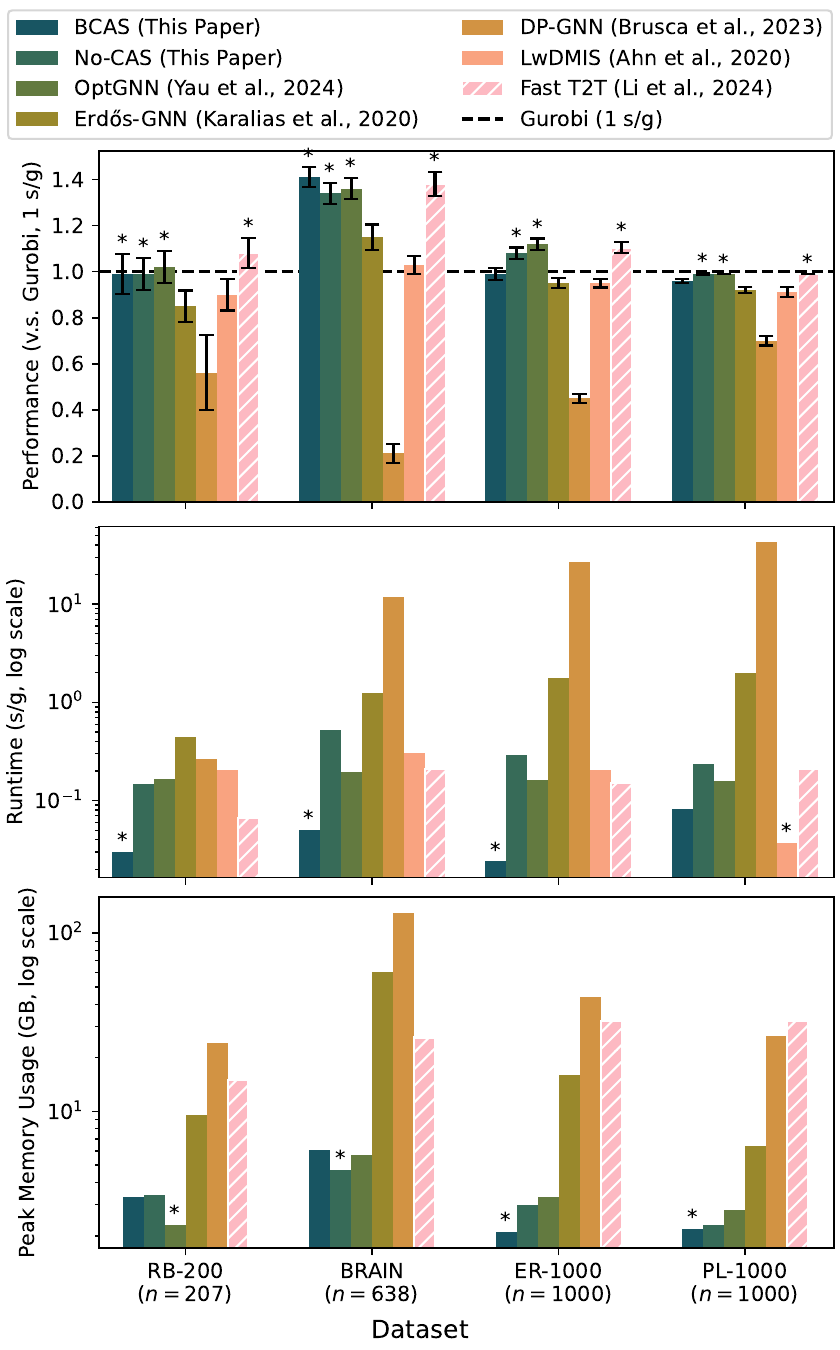}
    \caption{Models' performance (top), runtime (middle), and peak memory usage (bottom) across four datasets.
    The best method for each metric and dataset is marked with a ($\ast$); for performance, all methods whose performance ranges (error bars) are statistically indistinguishable from the method with the highest mean performance are considered ``best''.
    DP-GNN is very slow on the larger graphs (BRAIN, ER-1000, and PL-1000), so its results only reflect evaluation on these graphs' first \numprint{5000} (20\%) testing time steps.
    Also, LwDMIS failed to train on some graph snapshots as its search space collapsed to null; thus, we only report peak memory usage for methods that completed the entire training process.}
    \label{fig:results}
\end{figure}

\figtext~\ref{fig:results} shows that our models match the performance of the state-of-the-art methods---both unsupervised (OptGNN) and supervised (Fast T2T)---with competitive or superior runtimes, depending on the dataset.
Specifically, across all datasets, one or both of our BCAS and No-CAS variants achieve performance that is statistically indistinguishable from those of OptGNN and Fast T2T.
When directly comparing mean performance ratios, BCAS's are 0.88--1.04x of OptGNN's and 0.90--1.02x of Fast T2T's; analogously, No-CAS's are 0.96--1.00x of OptGNN's and 0.91--1.00x of Fast T2T's.
Taking a broader view, all four of these methods perform as well or better than Gurobi in a fraction of the time; on BRAIN in particular, this ensemble achieves a remarkable 1.34--1.41x improvement over the mixed integer programming solver.
The same cannot be said of the earlier unsupervised methods (Erd\H{o}s-GNN, DP-GNN, and LwDMIS), which often do not even achieve parity with Gurobi.

Our BCAS variant consistently achieves superior runtimes over OptGNN and Fast T2T (1.91--6.70x), offering a fast alternative to the state-of-the-art without sacrificing performance.
Peak memory usage comparisons against OptGNN are more variable---sometimes better and sometimes worse across datasets---but our models are certainly more memory efficient than the supervised Fast T2T model (0.07--0.23x).
Taken together, these performance and resource comparisons demonstrate the strength of our unsupervised update learning method as a viable alternative to repeatedly applying learning models for static graphs to individual graph snapshots.

\subsubsection*{Generalization}

We evaluated each model's generalization ability using dynamic graphs with 100x more nodes than those used for training (Table~\ref{tab:generalization}).\footnote{Specifically, we train each model on ER and PL dynamic graphs with \numprint{100} nodes and \numprint{35000} time steps and test them on ER and PL dynamic graphs with \numprint{10000} nodes and \numprint{5000} time steps.}
Our No-CAS variant significantly outperforms the earlier unsupervised methods (1.05--1.18x better than Erd\H{o}s-GNN, LwDMIS, and DP-GNN)---as does BCAS on the PL-10k dataset specifically---though neither are as fast as LwDMIS.
OptGNN is better still, achieving similar generalized performance to No-CAS while running 1.53x faster than LwDMIS.
The supervised method, Fast T2T, has the best generalization performance but middling runtimes (slower than BCAS, LwDMIS, and OptGNN, but faster than the rest).
On the negative side, it is somewhat surprising how poorly BCAS generalizes on the ER-10k dataset given its strong performance on ER-1000 (\figtext~\ref{fig:results}).
This may be caused by a mismatch of training vs.\ testing hyperparameters: in BCAS, both $\alpha$ and $\beta$ are fractions of the initial graph snapshot's diameter, which generalization inflates when considering 100x more nodes.

\begin{table}[tb]
    \centering
    \caption{Generalization results reported as mean $\pm$ std.dev.\ performance ratios vs. 10~s Gurobi runs followed by mean runtime per graph snapshot in parentheses.
    Notably, DP-GNN is prohibitively slow on these large graphs (taking nearly a week just to evaluate the first six snapshots of each dataset), so we omit its performance results.}
    \label{tab:generalization}
    \begin{tabular}{ccc}
        \toprule
        Dataset ($\rightarrow$) & ER-10k & PL-10k \\
        Method ($\downarrow$) & ($n = \numprint{10000}$) & ($n = \numprint{10000}$) \\
        \midrule
        \multirow{2}{*}{BCAS} & 0.63 $\pm$ 0.085 & \textbf{0.99 $\pm$ 0.002} \\
        & \textit{(0.015 s/g)} & (0.679 s/g) \\
        \multirow{2}{*}{No-CAS} & \textbf{1.17 $\pm$ 0.024} & \textbf{0.99 $\pm$ 0.001} \\
        & (5.349 s/g) & (3.372 s/g) \\
        \midrule
        \multirow{2}{*}{OptGNN} & 1.12 $\pm$ 0.012 & \textbf{0.99 $\pm$ 0.001} \\
        & (0.353 s/g) & \textit{(0.194 s/g)} \\
        \multirow{2}{*}{Erd\H{o}s-GNN} & 1.07 $\pm$ 0.023 & 0.94 $\pm$ 0.009 \\
        & (19.202 s/g) & (18.415 s/g) \\
        \multirow{2}{*}{DP-GNN} & N/A & N/A \\
        & (\numprint{36239.508} s/g) & (\numprint{39994.408} s/g) \\
        \multirow{2}{*}{LwDMIS} & 0.99 $\pm$ 0.020 & 0.92 $\pm$ 0.003 \\
        & (0.541 s/g) & (0.296 s/g) \\
        \midrule
        \multirow{2}{*}{\textcolor{gray}{Fast T2T}} & \textcolor{gray}{\textbf{1.27 $\pm$ 0.006}} & \textcolor{gray}{\textbf{0.99 $\pm$ 0.000}} \\
        & \textcolor{gray}{(1.29 s/g)} & \textcolor{gray}{(1.84 s/g)} \\
        \bottomrule
    \end{tabular}
\end{table}

\section{Conclusion} \label{sec:conclusion}

We present a single-shot, unsupervised learning model for the \textsc{Maximum-Independent-Set} (MaxIS) problem in dynamic graphs, where the topology evolves over time.
Using a simple GNN backbone, our model achieves results competitive with both supervised (Fast T2T~\cite{Li2024-fastt2t}) and unsupervised (OptGNN~\cite{Yau2024-aregraph}) state-of-the-art methods, breaking fertile ground for exploration with advanced architectures, representations, and objective design.
Unlike heuristic learning approaches, our method learns an update mechanism analogous to those used in traditional update algorithms for dynamic combinatorial optimization.
Our experiments across various dynamic graph topologies demonstrate our model's competitive performance against the state-of-the-art, surpassing earlier unsupervised methods like DP-GNN~\cite{Brusca2023-maximumindependent}, Erd\H{o}s-GNN~\cite{Karalias2020-erdosgoes} and LwDMIS~\cite{Ahn2020-learningwhat}.
%Motivated by the well-known relationships between Maximum Independent Set, Minimum Vertex Cover, and Maximum Clique, the strong performance of our model on dynamic MaxIS indicates its potential applicability to dynamic variants of Minimum Vertex Cover and Maximum Clique. 
In future work, we plan to extend our approach into a general framework for learning update algorithms applicable to a broader class of combinatorial optimization problems, leveraging recent advances in neural models.
%Moreover, a systematic exploration of alternative architectures and their relative advantages and limitations could further help us improve model performance. Finally, another prospect for performance gains lies in greedily collecting edge events occurring in isolated, non-overlapping $\alpha$- and $\beta$-neighborhoods and processing them in parallel as batches, thereby improving scalability. We leave the investigation of such batch update strategies to future work.

\subsubsection*{Limitations}

Although a thorough comparison to traditional, non-learning-based algorithms is beyond the scope of this paper, recent advances in update algorithms for MaxIS~\cite{Zheng2019-computingnearmaximum,Gao2022-dynamicapproximate} and past learning vs.\ non-learning comparisons in the static setting~\cite{Angelini2022-moderngraph} invite some reflection on this point.
We evaluated a simple greedy algorithm for MaxIS on our four dynamic graph topologies and found that it also achieves performance statistically indistinguishable from state-of-the-art, matching the claims we make about our own model.
For RB-200 in particular, the greedy algorithm runs 6.00x faster than the fastest learning model for this dataset (BCAS).
Thus, even with these preliminary investigations, there is an evident gap to close between learning-based approaches to combinatorial optimization and their traditional counterparts.

There are various aspects of our approach that should be investigated further.
The BCAS and No-CAS variants exemplify two specific instantiations of our model in terms of $\alpha$ and $\beta$ hyperparameters, but a broader characterization of model behavior as a function of these propagation radii may reveal further performance and runtime improvements.
Also, the dynamics for all our datasets sample only one edge event per time, albeit in a degree distribution-preserving manner.
It would be interesting to evaluate our model on large dynamic graphs whose structure and dynamics are both empirical.
Finally, our model makes limited use of temporal dynamics when learning its update mechanism.
Leveraging this rich source of information using recent advances in neural architectures and attention mechanisms is an important direction for future work.

\section*{Acknowledgments \& Code Availability}

This work is supported in part by NSF award CCF-2312537.
Source code for our method and complete reproducibility instructions for dataset processing and experiments is available at \url{https://github.com/DaymudeLab/LearnDynamicMaxIS}.

% Bibliography.
\bibliographystyle{abbrvnat}
\bibliography{ref}

\appendix

\subsection{Model Pseudocode} \label{app:pseudocode}

Algorithms~\ref{alg:initmaxis}--\ref{alg:intsolution} show pseudocode for our model's initial MaxIS generation procedure, update training process, and integral solution generation procedure, respectively.
Throughout, graph snapshots $G_t$, edge events $\mathcal{E}_t$, signals $s_t(\cdot)$, node memories $m_t(\cdot)$, node estimates $p_t(\cdot)$, and node losses $\ell_t(\cdot)$ are defined as they were in the main text.
Bold versions of these notations refer to the vector of all node properties of that type (e.g., $\mathbf{m}_t$ refers to all node memories).
In a slight abuse of notation, we use $\mathcal{N}_t^x(\mathcal{E}_t)$ to denote all nodes within an $x$ hop-distance from the edge event $\mathcal{E}_t$; i.e., the endpoints of the edge event are in $\mathcal{N}_t^0(\mathcal{E}_t)$, those endpoints and their neighbors are in $\mathcal{N}_t^1(\mathcal{E}_t)$, and so on.

\begin{algorithm}[ht]
\caption{Initial MaxIS Generation Epoch} \label{alg:initmaxis}
\begin{algorithmic}[1]
    \State Initialize all node memories as $\mathbf{m}_0 \gets \mathbf{0}$.
    \ForAll{edges $(u, v) \in E_0$}
        \State Let $\mathcal{E}$ be the edge event of adding $(u, v)$.
        \State Do event handling with signal $s \gets [\texttt{enc}(\mathcal{E})~||~0]$ as in Eq.~\ref{eq:signal}.
        \State Update $m_0(u) \gets \texttt{GRU}([m_0(u)~||~s])$ and $m_0(v) \gets \texttt{GRU}([m_0(v)~||~s])$ as in Eq.~\ref{eq:memory}.
    \EndFor
    \State Compute node estimates $\mathbf{p}_0$ from the updated node memories $\mathbf{m}_0$ as in Eqs.~\ref{eq:aggnbrs}--\ref{eq:estimate}.
    \State Compute node losses $\mathbf{\ell}_0$ from $\mathbf{p}_0$ as in Eq.~\ref{eq:loss}.
    \State Compute cumulative loss $L_0 \gets \sum_{v \in V} \ell_0(v)$.
\end{algorithmic}
\end{algorithm}

\begin{algorithm}[ht]
\caption{Update Training Epoch} \label{alg:training}
\begin{algorithmic}[1]
    \State Initialize node memories $\mathbf{m}_0$ using Algorithm~\ref{alg:initmaxis}.
    \ForAll{edge events $\mathcal{E}_t \in (\mathcal{E}_1, \mathcal{E}_2, \ldots, \mathcal{E}_T)$ in the training set}
        \ForAll{nodes $v \in \mathcal{N}_t^\alpha(\mathcal{E}_t)$}
            \State Do event handling with $s_t(v) \gets [\texttt{enc}(\mathcal{E}_t)~||~r_t(v)]$ as in Eq.~\ref{eq:signal}.
            \State Update $m_t(v)$ using $s_t(v)$ as in Eq.~\ref{eq:memory}.
        \EndFor
        \ForAll{nodes $v \in \mathcal{N}_t^\beta(\mathcal{E}_t)$}
            \State Compute estimate $p_t(v)$ from $m_t(v)$ as in Eqs.~\ref{eq:aggnbrs}--\ref{eq:estimate}.
        \EndFor
        \ForAll{nodes $v \in \mathcal{N}_t^\beta(\mathcal{E}_t)$}
            \State Compute node loss $\ell_t(v)$ from $\mathbf{p}_t$ as in Eq.~\ref{eq:loss}.
        \EndFor
        \State Compute cumulative loss $L_t \gets \sum_{v \in \mathcal{N}_t^\beta(\mathcal{E}_t)} \ell_t(v)$.
    \EndFor
\end{algorithmic}
\end{algorithm}

\begin{algorithm}[ht]
\caption{Integral Solution Generation (Testing \& Inference)} \label{alg:intsolution}
\begin{algorithmic}[1]
    \State Initialize node memories $\mathbf{m}_0$ using Algorithm~\ref{alg:initmaxis}.
    \ForAll{edge events $\mathcal{E}_t \in (\mathcal{E}_1, \mathcal{E}_2, \ldots, \mathcal{E}_T)$ in the testing set}
        \ForAll{nodes $v \in \mathcal{N}_t^\alpha(\mathcal{E}_t)$}
            \State Do event handling with $s_t(v) \gets [\texttt{enc}(\mathcal{E}_t)~||~r_t(v)]$ as in Eq.~\ref{eq:signal}.
            \State Update $m_t(v)$ using $s_t(v)$ as in Eq.~\ref{eq:memory}.
        \EndFor
        \ForAll{nodes $v \in \mathcal{N}_t^\beta(\mathcal{E}_t)$}
            \State Compute estimate $p_t(v)$ from $m_t(v)$ as in Eqs.~\ref{eq:aggnbrs}--\ref{eq:estimate}.
        \EndFor
        \ForAll{nodes $v \in \mathcal{N}_t^\beta(\mathcal{E}_t)$} \Comment{Estimate Rounding}
            \If {$p_t(v) \geq 0.5$} $I_t(v) \gets 1$.
            \Else {} $I_t(v) \gets 0$.
            \EndIf
        \EndFor
        \While {there exists a node $v \in \mathcal{N}_t^\beta(\mathcal{E}_t)$ with violations} \Comment{Violation Removal}
            \State Choose the node $v \in \mathcal{N}_t^\beta(\mathcal{E}_t)$ with the most violations.
            \State Remove $v$ from the MaxIS wtih $I_t(v) \gets 0$.
        \EndWhile
    \EndFor
\end{algorithmic}
\end{algorithm}

\subsection{Estimate Threshold Sensitivity} \label{app:estimates}

\begin{figure}[ht]
    \centering
    \includegraphics[width=\columnwidth]{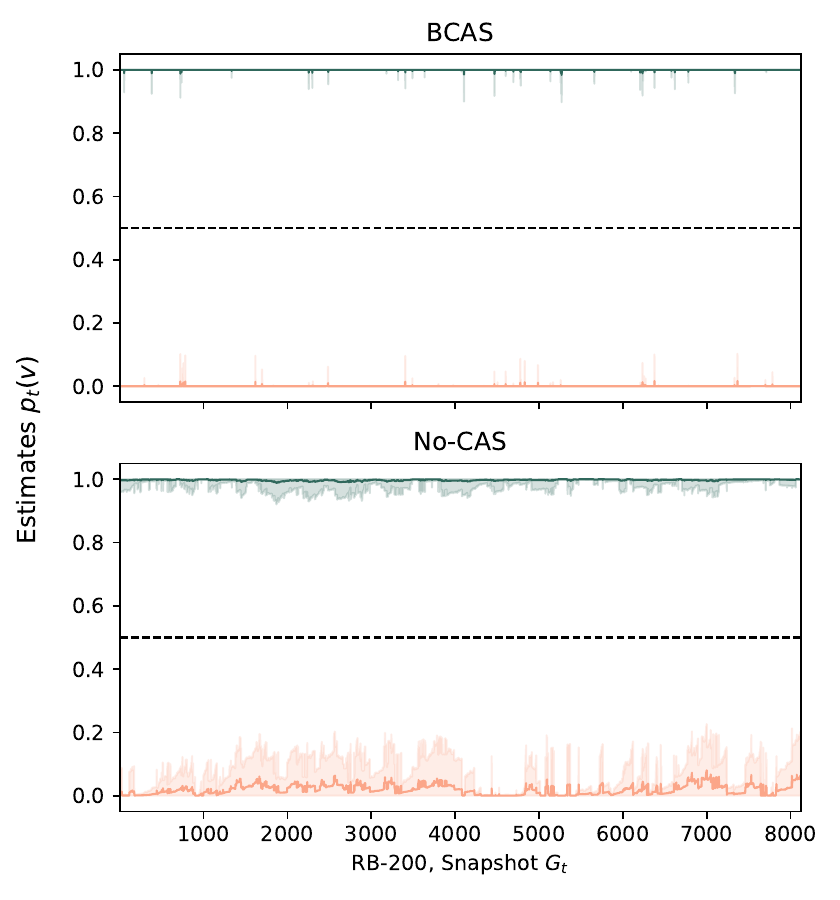}
    \caption{Node estimates $p_t(v)$ of membership in the candidate solution across testing snapshots of the RB-200 dataset.
    After partitioning these estimates by those above (green) and below (orange) the 0.5 threshold (dashed line), it is clear from each part's mean estimate (solid line) and standard deviation (shaded region) that all estimates are nearly-one or nearly-zero.}
    \label{fig:estimates}
\end{figure}

When generating an integral MaxIS solution from node estimates $p_t(v)$, our model initially includes all nodes with estimates $p_t(v) \geq 0.5$ in the candidate solution.
\figtext~\ref{fig:estimates} shows that the end results are not particularly sensitive to this choice of threshold: almost all estimates are nearly-zero or nearly-one, so thresholds as high as 0.75 and as low as 0.25 would yield nearly identical results.

\end{document}